\documentclass[sigconf]{acmart}

\AtBeginDocument{%
  }

\usepackage{balance}
\begin{document}
\settopmatter{printacmref=false}
\setcopyright{none}
\renewcommand\footnotetextcopyrightpermission[1]{}
\pagestyle{plain}
\copyrightyear{2024}
\acmYear{2024}
\setcopyright{acmlicensed}\acmConference[ICMR '24]{Proceedings of the 2024 International Conference on Multimedia Retrieval}{June 10--14, 2024}{Phuket, Thailand}
\acmBooktitle{Proceedings of the 2024 International Conference on Multimedia Retrieval (ICMR '24), June 10--14, 2024, Phuket, Thailand}
\acmDOI{10.1145/3652583.3658036}
\acmISBN{979-8-4007-0619-6/24/06}

\title{RGB-D Video Object Segmentation via Enhanced Multi-store Feature Memory}

\author{Boyue Xu}
\affiliation{%
  \institution{State Key Laboratory for Novel Software Technology, Nanjing University}
  \city{Nanjing}
  \country{China}
}
\email{xuby@smail.nju.edu.cn}

\author{Ruichao Hou}
\authornote{Corresponding author.}
\affiliation{%
  \institution{State Key Laboratory for Novel Software Technology, Nanjing University}
  \city{Nanjing}
  \country{China}
}
\email{rc\_hou@smail.nju.edu.cn}

\author{Tongwei Ren}
\affiliation{%
  \institution{State Key Laboratory for Novel Software Technology, Nanjing University}
  \city{Nanjing}
  \country{China}
}
\email{rentw@nju.edu.cn}

\author{Gangshan Wu}
\affiliation{%
  \institution{State Key Laboratory for Novel Software Technology, Nanjing University}
  \city{Nanjing}
  \country{China}
}
\email{gswu@nju.edu.cn}
\renewcommand{\shortauthors}{Boyue Xu, Ruichao Hou, Tongwei Ren, \& Gangshan Wu}

\begin{abstract}
The RGB-Depth (RGB-D) Video Object Segmentation (VOS) aims to integrate the fine-grained texture information of RGB with the spatial geometric clues of depth modality, boosting the performance of segmentation. However, off-the-shelf RGB-D segmentation methods fail to fully explore cross-modal information and suffer from object drift during long-term prediction. In this paper, we propose a novel RGB-D VOS method via multi-store feature memory for robust segmentation. Specifically, we design the hierarchical modality selection and fusion, which adaptively combines features from both modalities. Additionally, we develop a segmentation refinement module that effectively utilizes the Segmentation Anything Model (SAM) to refine the segmentation mask, ensuring more reliable results as memory to guide subsequent segmentation tasks. By leveraging spatio-temporal embedding and modality embedding, mixed prompts and fused images are fed into SAM to unleash its potential in RGB-D VOS. Experimental results show that the proposed method achieves state-of-the-art performance on the latest RGB-D VOS benchmark. 
\end{abstract}

\ccsdesc[500]{Computing methodologies~Video segmentation}
\ccsdesc[300]{Computing methodologies~Artificial intelligence}
\ccsdesc[300]{Computing methodologies~Computer vision}

\keywords{RGB-Depth, Video Object Segmentation, Memory Mechanism, Segment Anything Model}

\hyphenpenalty=8000
\tolerance=1000
\maketitle
\section{Introduction}
Video object segmentation (VOS)~\cite{VOS} aims to continuously segment specific object masks given in the first frame throughout the video sequence, which has numerous applications in autonomous driving~\cite{driving,drive,drive2}, 3D reconstruction~\cite{reconst,reconst2,reconst3}, surveillance~\cite{xu1,tjh}.

\begin{figure}[t]
    \centering
    \includegraphics[scale=0.53]{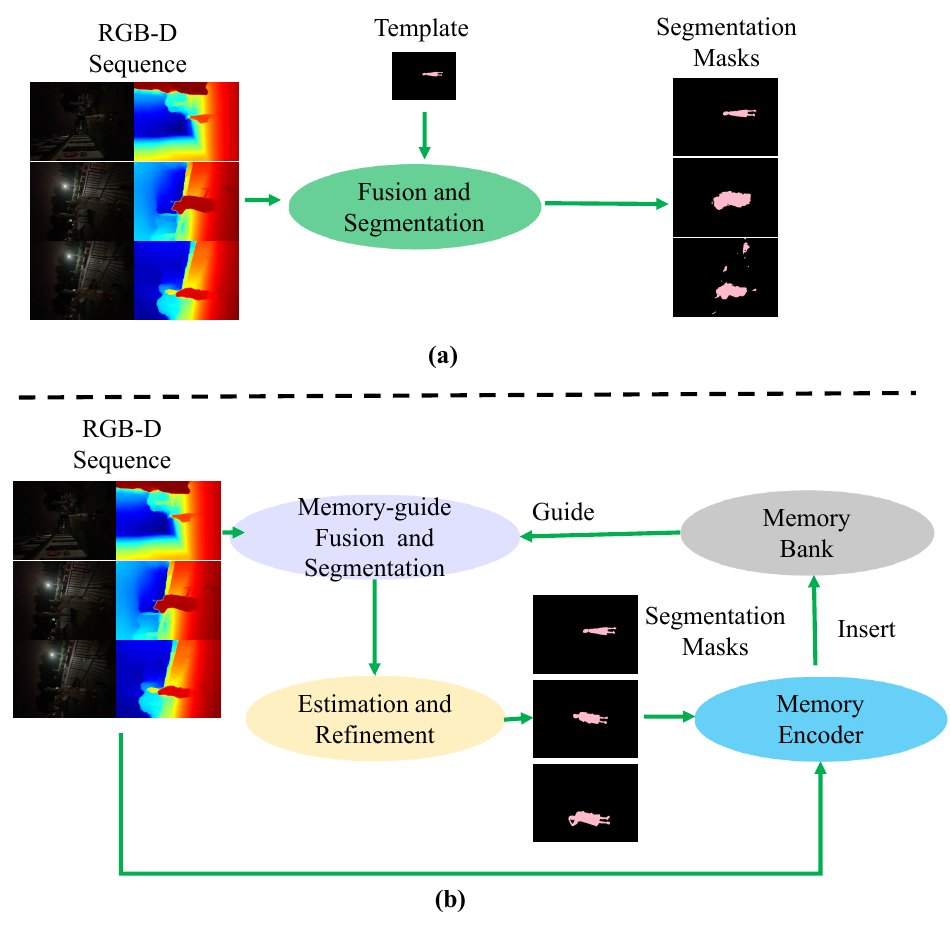}
    \caption{Comparison of the framework between different RGB-D VOS methods. (a) RGB-D VOS methods without memory, which use template to guide fusion and segmentation. (b) The proposed method which use memory to guide fusion and segmentation.}
    \label{fig:intro}
\end{figure}

RGB VOS methods face various challenges in scenarios such as extreme illumination, complex backgrounds, and occlusion. To address these issues, RGB-Depth (RGB-D) VOS~\cite{arkittrack} introduces depth modality which offers additional spatial geometric clues for more robust segmentation. However, as shown in Figure~\ref{fig:intro}(a), RGB-D VOS methods which use template to guide fusion and segmentation may encounter challenges in complex scenarios and initial template may not guide the subsequent segmentation in long-term videos, decreasing the robustness of segmentation.

Memory mechanism is used to address these issues in RGB VOS~\cite{Xmem,STCN,DEAOT,AOT,hu2021learning,QDMN}, 
which mine the spatio-temporal information and appearance features from previous segmentation results to guide subsequent segmentation. Compared to traditional approaches, memory mechanism effectively addresses the problem of poor feature representation caused by continuous propagation relying solely on adjacent frames.
Among all memory-based methods, XMem~\cite{Xmem} achieves satisfactory results by constructing a multi-store memory network inspired by Atkinson-Shiffrin memory model ~\cite{memory}. It consists of sensory memory, working memory and long-term memory that significantly improves the robustness of segmentation through effective insertion of memory content. XMem effectively inserts memory content, significantly improving the robustness of segmentation. 

Inspired by XMem, we propose an enhanced multi-store memory network for RGB-D VOS. However, there are two challenges that need to be addressed: (1) How to adaptively fuse complementary information from RGB-D modalities? (2) How to estimate the quality and reliability of inserted segmentation results as guidance for accurate subsequent segmentation?

As shown in Figure~\ref{fig:intro}{(b), the hierarchical modality selection and fusion (HMSF) is proposed to fuse RGB-D features under the guide of memory. HMSF is capable of extracting the complementary features from both modalities and selectively fusing multi-modal features, making it suitable for RGB-D feature fusion during segmentation and content encoding in a multi-store memory network. 
To enhance the reliability of segmentation results, we attempt to leverage the powerful segmentation capabilities of the Segment Anything Model (SAM)~\cite{SAM} for refinement. To the best of our knowledge, we are the first to introduce SAM into the RGB-D VOS. Specifically, we take the fused image and the mixed prompt as the input of SAM. This paper focuses on generating pixel-wise RGB-D fused images and effective prompts to reduce the false positive pixels or regions, thereby guiding more accurate segmentation.
To this end, we propose spatio-temporal embedding and modality embedding to further improve performance. The spatio-temporal embedding integrates historical information from the entire video segmentation process to provide prompts for SAM, while modality embedding effectively combines RGB-D images through a pixel-wise fusion strategy. The fused weights are calculated by the quality of modality from HMSF and the significance of the region to facilitate the segmentation performance and refine the memory content.
Experimental results demonstrate that the proposed method achieves the best results on the latest RGB-D VOS dataset. 

The main contributions of this paper are summarized as follows:

\begin{itemize}
  \item We propose a novel RGB-D VOS method based on multi-store feature memory, enabling robust segmentation of RGB-D sequences in diverse and complex scenarios.
  \item We present the HMSF to effectively fuse hierarchical and cross-modality features. Also, it can facilitate the encoding of memory.
  \item We develop a segmentation refinement module that incorporates spatio-temporal embedding and modality embedding with SAM, significantly enhancing the reliability of segmentation results.

\end{itemize}
\section{Related Work}
\begin{figure*}[t]
    \centering
    \includegraphics[scale=0.5]{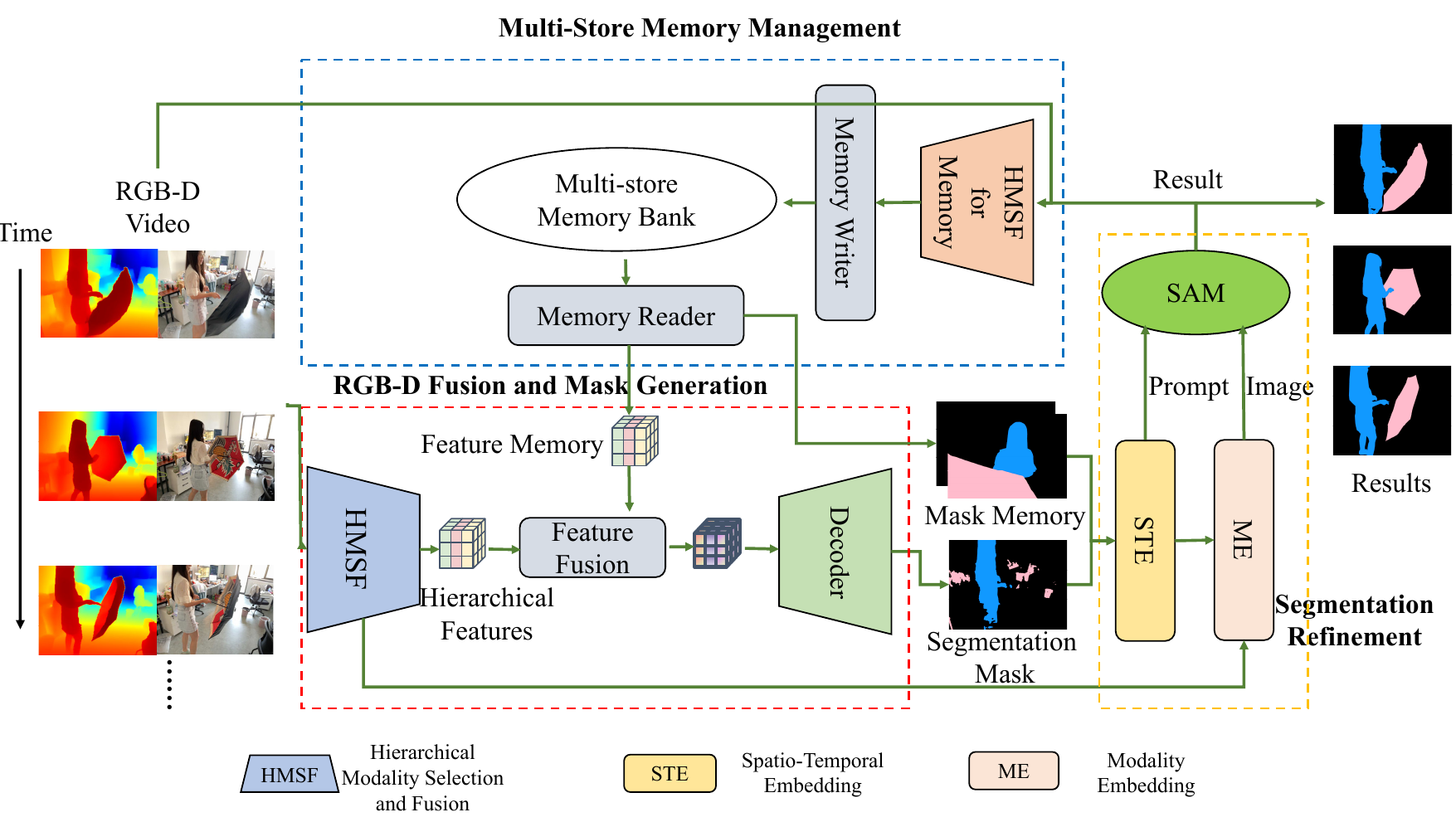}
    \caption{The framework of the proposed method, consists of RGB-D fusion and mask generation, segmentation refinement and multi-store memory management.}
    \label{fig:pipeline}
\end{figure*} 
\subsection{Semi-supervised VOS}

Video object segmentation encompasses unsupervised VOS~\cite{un1,un2}, semi-supervised VOS~\cite{QDMN,RPCM,Xmem,DEAOT}, and interactive VOS~\cite{active1,active2}. Semi-supervised VOS annotates the object masks in the first frame to guide the segmentation of the rest frames.
The semi-supervised VOS methods can be categorized into motion-aware VOS methods and detection-aware VOS methods.

Motion-aware VOS methods utilize optical flow or CNN-based learning for mask refinement. Optical flow-based methods employ motion cues to estimate pixel changes over time~\cite{lucid,segflow}, while CNN-based learning methods refine the object mask frame-to-frame by leveraging temporal information~\cite{time1} or combining RNNs~\cite{youtube} with current mask predictions. Although these methods show promising performance, they still lack robustness when dealing with challenging attributes, \emph{i.e.}, occlusion or fast motion in long-term videos.

Detection-aware VOS methods primarily involve learning an appearance model for pixel-wise object detection and segmentation. For instance, 
Caelles \emph{et al.}~\cite{caelles2017one} employ fully convolutional neural networks on static images to segment objects by fine-tuning the first frame of the video sequence.
Some other methods~\cite{shin2017pixel,hu2018videomatch} employ pixel-matching techniques to segment objects based on feature matching with a template.
Yang \emph{et al.}~\cite{AOT,DEAOT} embed multiple objects into the same embedding space, then uniformly and simultaneously propagate all object embedding. 
Xmem~\cite{Xmem}, inspired by human memory mechanisms, offers an innovative propagation method, avoiding the out-of-memory crash.
These methods perform well in handling occlusion or fast motion scenarios. However, distinguishing cross-temporal similar objects remains challenging. 

In RGB-D VOS, Zhao \emph{et al.}~\cite{arkittrack} pioneer the emerging area by introducing the first benchmark known as ARKitTrack, which opens up a new frontier. However, there is room for decreasing the complexity of the fusion model and improving its performance in long-term segmentation.
We propose a multi-store feature memory for robust RGB-D segmentation.
\subsection{RGB-D Video Object Tracking}
As RGB-D VOS is a relatively new domain with limited prior work, it is beneficial to provide an introduction to the related field of RGB-D tracking~\cite{dal,rgbd1k,protrack,VIPT,xu1}, which shares a similar processing scheme.

RGB-D tracking is a type of multi-modality object tracking\cite{hou1,hou2,RGBT2,rgbt1,UAV}.
Song \emph{et al.}~\cite{PTB} establish the first RGB-D tracking dataset, pioneering the task of RGB-D tracking. Zhong \emph{et al.}~\cite{zhongRGBD} introduce a method for extracting key points from dense depth maps for use in RGB-D tracking. 
Bibi \emph{et al.}\cite{bibi} employ appearance models and 3D spatial motion models to estimate object positions.
Kart \emph{et al.}~\cite{KARTrgbd} integrate discriminative correlation filters (DCF) to propose a general framework for RGB-D tracking. 
Qian \emph{et al.}~\cite{dal} address the occlusion challenge by embedding depth into deep features and training discriminators.
Zhao \emph{et al.}~\cite{tsdm} employ depth to generate object masks and accurately cut out prediction results.
Yan \emph{et al.}~\cite{DepthTrack} explore pseudo-color images derived from depth data and investigate the integration with RGB images. With the advancement of transformer, Zhu \emph{et al.}~\cite{rgbd1k} incorporate transformers to fuse multi-modal features.
Moreover, some researchers~\cite{protrack,VIPT} apply prompt learning to adapt existing RGB trackers for RGB-D tracking. 

\section{METHODOLOGY}
\subsection{Network Architecture}

As illustrated in Figure~\ref{fig:pipeline}, the proposed method  comprises three modules: RGB-D fusion and mask generation module, segmentation refinement module, and multi-store memory management module. Specifically, the RGB-D fusion and mask generation module aims to fuse RGB-D dual-modality features and integrate them with multi-store feature memory to produce segmentation results. The segmentation refinement module flexibly utilizes the SAM~\cite{SAM} to refine segmentation results and ensure more accurate results as memory to guide subsequent segmentation. The multi-store memory management module encodes and stores both RGB-D images and segmentation results as feature memory.

The proposed method utilizes dual-modality image sequences as input, which are initially processed by the HMSF for feature extraction and fusion. The output from HMFS is then combined with the feature memory in the multi-store memory bank, followed by decoding to obtain accurate segmentation results. To further improve the segmentation performance and guide the subsequent segmentation, we incorporate SAM~\cite{SAM} to refine segmentation masks. In particular, we estimate the reliability of segmentation masks and generate mixed prompts for SAM to maintain the consistency of the given prompts via spatio-temporal embedding. Next, we fuse the RGB-D images using a pixel-wise strategy by using modality embedding. Subsequently, the refined masks are encoded via the HMSF for memory and inserted into the multi-store memory bank that captures historical information of segmentations, thus preventing error accumulation in long-term videos. 

\subsection{RGB-D Fusion and Mask Generation}
To extract and integrate hierarchical complementary features from different modalities, we propose hierarchical modality selection and fusion. As shown in Figure~\ref{fig:hmsf}(a), the HMSF firstly extracts features of different modalities using ResNet50~\cite{resnet}. Then it adaptively fuses dual-modality hierarchical features across layer 1 to layer 3 of ResNet50 via modality selection and fusion, which combines the shallow features and deep features to enhance the multi-modality feature representation. The fused hierarchical features $F_4$, $F_8$, and $F_{16}$ are utilized for segmentation. 
\begin{figure}[t]
    \centering
    \includegraphics[scale=0.6]{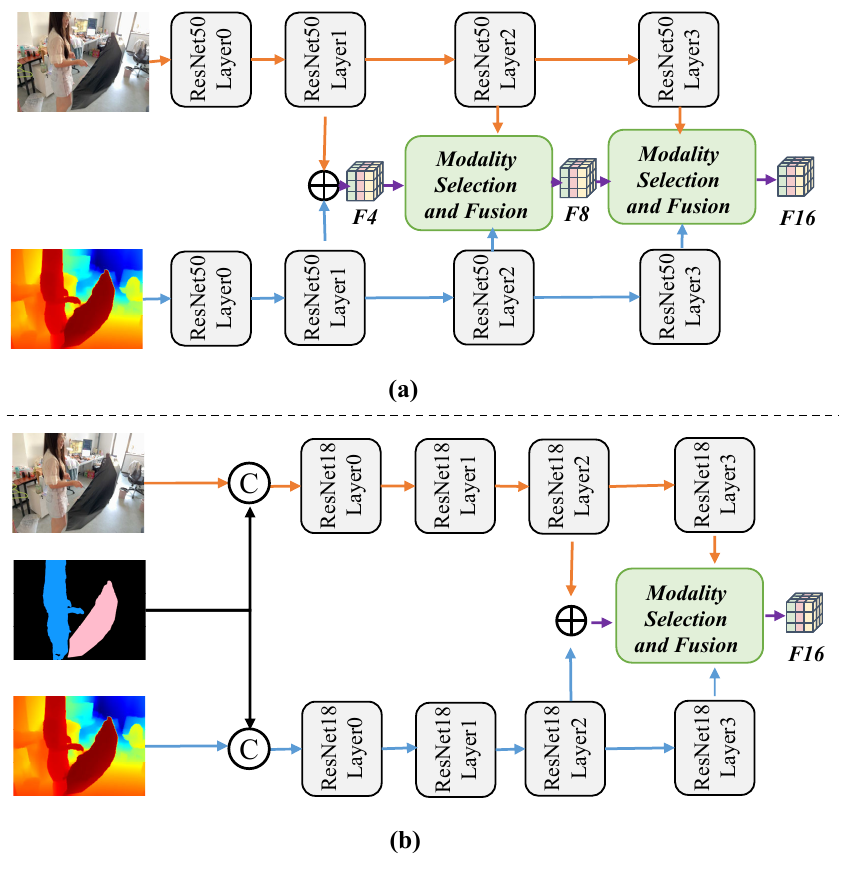}
    \caption{The detail of hierarchical modality selection and fusion.(a) Hierarchical modality selection and fusion used in RGB-D fusion and mask generation module. (b) Hierarchical modality selection and fusion for memory used in multi-store memory management module. }
    \label{fig:hmsf}
\end{figure} 
\begin{figure}[t]
    \centering
    \includegraphics[scale=0.63]{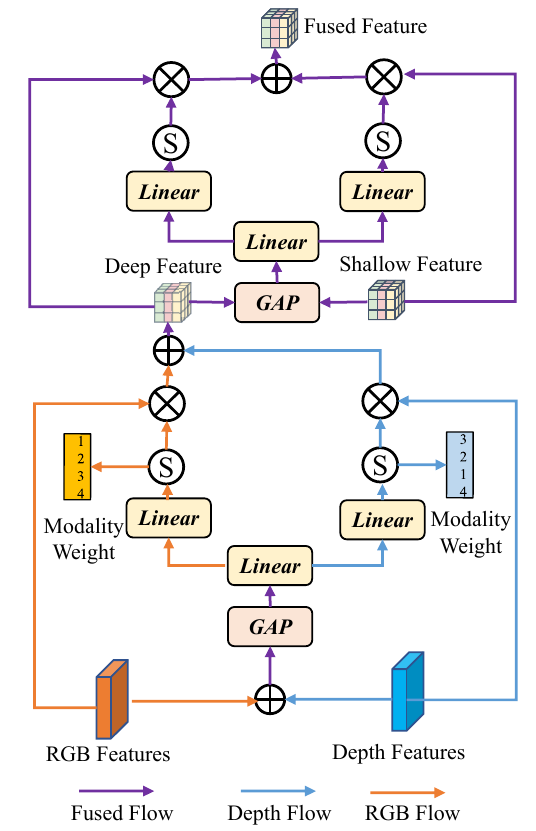}
    \caption{The details of modality selection and fusion. }
    \label{fig:msf}
\end{figure} 
\begin{figure}[t]
    \centering
    \includegraphics[scale=0.52]{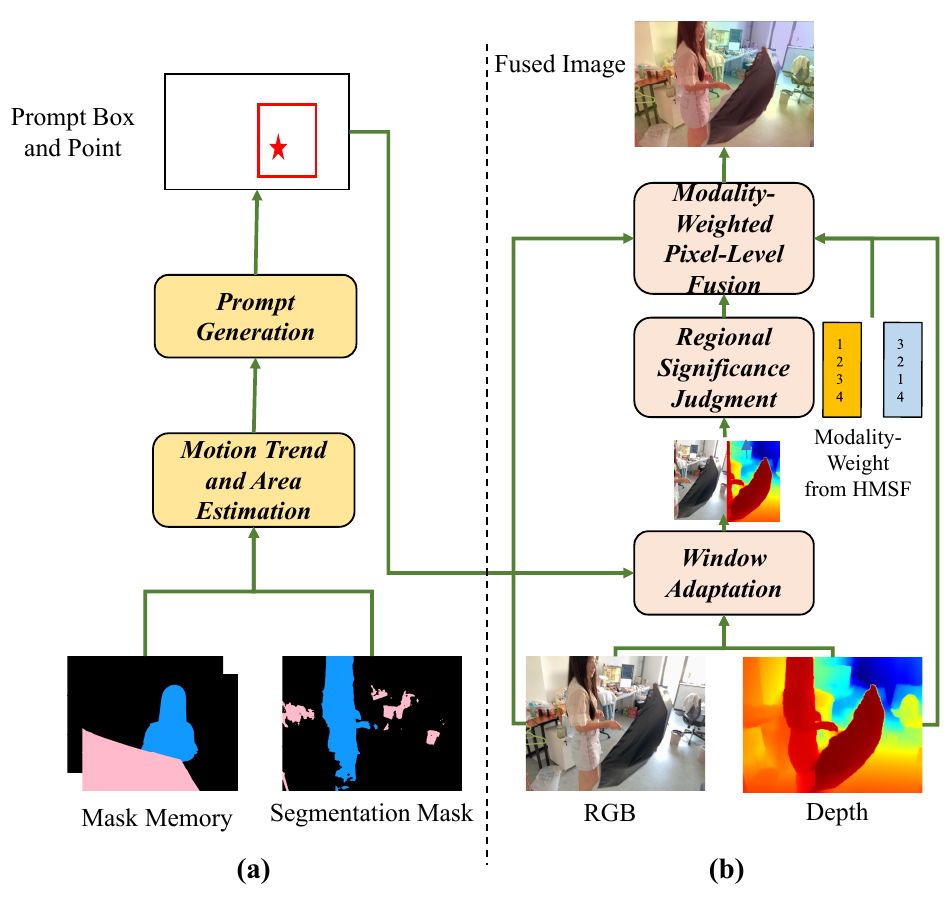}
    \caption{The details of segmentation refinement. (a) The details of spatio-temporal embedding, which generates mixed prompts. (b) The details of modality embedding, which generates fused images.}
    \label{fig:embed}
\end{figure} 

The details of modality selection and fusion are shown in Figure~\ref{fig:msf}. Initially, the RGB and depth features from the current layer undergo global average pooling to extract global features. These global features are then distributed through the RGB and depth flows to generate channel-level modality weights, which are crucial for selectively fusing dual-modality features and guiding pixel-wise fusion within the segmentation refinement module. Next, the modality weights are multiplied by the corresponding feature channels to obtain discriminative features. Deep features are derived by summing the above discriminative features, which is formulated as follows:
\begin{align}
F_{global} &= FC\left(GAP\left(F_{RGB} \oplus F_D\right)\right), \label{eq1} \\
\hat{W}_i &= \sigma\left(FC_i\left(F_{global}\right)\right), \quad i \in \{RGB, D\}, \label{eq2} \\
\hat{F}_i &= F_i \otimes W_i, \quad i \in \{RGB, D\}, \label{eq3} \\
F_{deep} &= \hat{F}_{RGB} \oplus \hat{F}_D, \label{eq4}
\end{align}
where $\operatorname{FC(\cdot)}$ is the linear layer; $\operatorname{GAP(\cdot)}$ represents global average pooling; $F_{RGB}$, $F_D$, $W_{RGB}$, $W_D$ represent RGB features, depth features, RGB weights, depth weight respectively; $\oplus$ denotes element-wise addition and $\otimes$ denotes the element-wise multiplication operation.

Subsequently, both shallow features and deep features are fused separately using the same way. Modality selection and fusion aim at fusing hierarchical features from multi-scale layers to expand the receptive field and thus enhance the representational power of the fused features.

Once obtaining the hierarchical features, we retrieve the best-matched features from the memory bank and merge them with the hierarchical features for decoding to generate segmentation masks. Specifically, the features stored in the memory are concatenated with $F_{16}$. Subsequently, these hierarchical features are decoded into a segmentation mask using a series of upsampling operations. The decoder and memory retrieval operations follow XMem~\cite{Xmem}.
\subsection{Segmentation Refinement}
To improve the reliability of segmentation results and guide subsequent segmentation, we propose a segmentation refinement module that flexibly leverages SAM. This module incorporates spatio-temporal embedding to estimate the quality of segmentation, generating mixed prompts for guiding SAM. Moreover, it employs modality embedding to evaluate the reliability of different modalities, thereby generating pixel-wise fused RGB-D images that are enriched with cross-modal information for SAM. 

\textbf{Spatio-Temporal Embedding.}
The spatio-temporal embedding leverages historical segmentation results to estimate the reliability of current segmentation and generate mixed prompts, which are composed of box prompt and point prompt. While the box-only prompt can designate the object area, it often struggles to accurately specify the object in complex backgrounds. On the other hand, the point-only prompt cannot precisely predict the scale of objects. To bridge this gap between these two types of prompts, the mixed prompt capitalizes on their respective strengths, unleashing SAM's potential for accurate segmentation.

As shown in Figure~\ref{fig:embed}(a), spatio-temporal embedding considers historical spatial trends of the objects, which involves motion trends and area estimation. Deviation from the expected motion trend in the current segmentation result indicates a potential error in object positioning. In such cases, the cluster center from the previous mask memory is used to generate the point prompt. Conversely, if the current position is deemed highly reliable, the cluster center of the current mask serves as the point prompt. Specifically, we define a threshold for significant object shift denoted as $M$. When the deviation between the current position and the historical trend exceeds $M$, it implies a potential error prompting the generation of point prompts using stored memory content. Conversely, if the deviation  falls within $M$, the point prompt is generated using the cluster center of the current position. Additionally, area estimation is performed, if there is a considerable disparity in mask area between the segmentation mask and mask memory, it may indicate unreliable segmentation and we generate the box prompt using the outer enclosing box of mask memory. Otherwise, the outer enclosing box of the current segmentation mask is used as box prompt. The mixed prompt consisting of both box and point information is then fed into SAM to provide guidance for modality embedding and generate reliable dual-modality images.

\textbf{Modality Embedding.}
The modality embedding explores the collaborative and heterogeneous nature of different modalities, integrating RGB-D complementary information and providing SAM with pixel-wise fused RGB-D images. In complex scenarios, the fused weights learned from the training set may yield suboptimal fused results. Therefore, we enhance the reliability of the fused weights by combining regional significance assessment with the modality weights of HMSF. 

As illustrated in Figure~\ref{fig:embed}(b),  
we first crop the dual-modality images to an appropriate size based on the box prompt from the spatio-temporal embedding to extract the region of interest and minimize excessive background interference. Subsequently, we estimate the significance of the depth image within the cropped region. If the significance rate is low, incorporating depth modality might introduce segmentation errors. Conversely, if the significance rate is high, we can employ the modality weights obtained from the HMSF for pixel-wise fusion. The regional significance assessment involves calculating the entropy of color distribution in a pseudo-color transformed version of the depth image, which can be calculated as follows:
\begin{equation}
E=-\sum_{i=0}^{255} \sum_{j=0}^{255} \sum_{k=0}^{255} H[i, j, k] \cdot \log _2(H[i, j, k]),
\end{equation}
where $E$ represents the value of entropy, with a higher value indicating a higher significance rate;  
$H[i,j,k]$ denotes the proportion of pixels in the color space with red, green, and blue channel values of $i$, 
$j$, and $k$, respectively. After estimating the significance of the region, we determine whether to fuse depth information. Subsequently, using the modality weights obtained from the HMSF, we perform a weighted element-wise addition of the dual-modality images. This aims to meet the input requirement of the SAM while also maintaining effective fusion of the modalities~\cite{fusion}.
Ultimately, this process provides the SAM with a fused image for segmentation refinement.
\subsection{Multi-Store Memory Management}

Memory mechanism in VOS facilitates the extraction of appearance features and establishment of spatio-temporal connections from previous segmentation results. To enhance the robustness of RGB-D VOS, we adopt the multi-store memory model introduced by XMem~\cite{Xmem}, modifying it to suit the requirements of dual-modalities. 

The multi-store memory model, inspired by Atkinson-Shiffrin memory model ~\cite{memory}, divides memory into sensory memory, working memory, and long-term memory. The network encodes images and segment results at distinct frequencies and stores them in each feature memory. During retrieval, a similarity matrix is computed based on the current frame's features to match relevant content from the multi-store memory.

To enhance the reliability of segmentation masks as memory by encoding complementary features of RGB-D images, we propose the HMSF for memory as the memory encoder. The proposed method utilizes modality selection and fusion to encode the refined masks from the segmentation refinement module along with RGB-D images as the memory content.
As shown in Figure~\ref{fig:hmsf}(b), we employ the ResNet18~\cite{resnet} as the backbone for encoder. The refined segmentation results are concatenated with RGB and depth images respectively, and the modality selection and fusion are used to extract features from layer 2 and layer 3. These fused features are stored in memory according to the following formula:
\begin{equation}
M=Fu((RGB_2\oplus D_2),RGB_3,D_3),
\end{equation}
where $Fu(\cdot)$ represents  modality selection and fusion; $M$ represents the memory features; $RGB_2$, $D_2$, $RGB_3$, and $D_3$ denote the second and third layer features of RGB and depth, respectively. This way allows the multi-store memory to adapt to the multi-modal memory content of RGB-D, effectively preventing errors in memory content that could mislead subsequent segmentation.
\subsection{Loss Function}
Following XMem~\cite{Xmem}, we employ bootstrapped cross-entropy loss and dice loss as loss functions~\cite{bce}. The bootstrapped cross-entropy loss can be calculated as:
\begin{equation}
\mathcal{L}_{bce}=\frac{1}{\left|S_l\right|} \sum_{m_i^l, g_i \in S_l}\left\{F\left(m_i^l\right)<\eta\right\} \mathbf{C}\left(g_i, F\left(m_i\right)\right),
\end{equation}
where $F(\cdot)$ is the output probability for a labeled example $m_i$ ; $g_i$ is ground truth, and $C(\cdot)$ represents the cross-entropy loss. The dice loss can be calculated as:
\begin{equation}
\mathcal{L}_d=1-\frac{2|m \cap g|}{|m|+|g|},
\end{equation}
where $m$ and $g$ represent predict mask and ground truth mask, respectively. The total loss is calculated as the sum of the bootstrapped cross-entropy loss and the dice loss:
\begin{equation}
\mathcal{L}_{total}=\mathcal{L}_{bce}+\mathcal{L}_{d}.
\end{equation}
\section{Experiments}
\subsection{Datasets and and Metrics}
The ARKitTrack~\cite{arkittrack} is the most recent RGB-D VOS dataset, which is utilized for comparative experiments against state-of-the-art methods. 
This dataset consists of more than 200 pairs of RGB-D sequences of $1920\times 1440$ resolution which collected in real-world scenarios. Each video sequence contains synchronized and aligned RGB frames and depth maps. The sequences encompass various challenging scenarios such as similar objects, occlusions, and extreme illumination.

We use $\mathcal{J_M}$ , $\mathcal{F_M}$, and $\mathcal{J\&F}$ measure~\cite{VOS} as evaluation metrics used in these experiments. Specifically, $\mathcal{J_M}$ denotes region similarity and is calculated as the intersection over union (IoU) between the predicted object segmentation mask and the ground truth. It can be calculated using the following formula:
\begin{equation}
\mathcal{J_M}=\frac{M \cap G}{M \cup G},
\end{equation}
where $M$ is the predicted mask, while 
$G$ denotes the ground truth. $\mathcal{F_M}$ stands for contour accuracy which is calculated based on the precision and recall of the contour. It can be calculated as follows:
\begin{equation}
\mathcal{F_M}=\frac{2 P_c R_c}{P_c+R_c},
\end{equation}
where $P_c$ represents precision and $R_c$ denotes recall. The metric $\mathcal{J\&F}$ is the average of $\mathcal{J_M}$ and $\mathcal{F_M}$:
\begin{equation}
\mathcal{J\&F}=\frac{\mathcal{J_M}+\mathcal{F_M}}{2}.
\end{equation}
\subsection{Implementation Details}
The proposed model is trained on a server equipped with a 5.2GHz CPU and four 3090 GPUs with a total of 96GB of memory. During training, we employ the AdamW optimizer~\cite{adam} with a learning rate set to $1e^{-5}$ and a batch size of 8. The model undergoes a total of 120K iterations on the training set. All other training parameters are consistent with the baseline~\cite{Xmem}. We utilize the ViT-H version weights of SAM~\cite{SAM}. Furthermore, the threshold for regional significance $E$ is set to 6, while the threshold for significant object shift $M$ is set to 500 pixels. The parameter amount of the proposed method without SAM is 64.9M, when the input images are paired 1920$\times$1440 RGB-D dual-modal images, the inference efficiency is about 1.5 FPS.

\subsection{Comparison with the State-of-the-Art}
\begin{table}[t]
\begin{center}
\caption{Comparison results of the proposed method against the competing methods on ARKitTrack test set. The upper section lists RGB methods, and the lower section includes RGB-D methods. The best results are highlighted in \textbf{bold}.} \label{tab:1}
\begin{tabular}{cccccc}
\hline
Methods      & Year  &  $\mathcal{J_M}\uparrow$  &  $\mathcal{F_M}\uparrow$   &  $\mathcal{J\&F}\uparrow$ \\ \hline
STCN~\cite{STCN}   & 2021  & 0.489 &0.560  & 0.525   \\
AOT~\cite{AOT} & 2021   & 0.555 &0.627  & 0.582   \\
RPCM~\cite{RPCM} & 2022   & 0.492 &0.527  & 0.509   \\
QDMN~\cite{QDMN} & 2022   & 0.276 &0.337  & 0.306   \\
XMem~\cite{Xmem} & 2022   & 0.541 &0.565  & 0.553   \\
 \hline
STCN\_RGBD~\cite{STCN}   & 2021  & 0.498 &0.574  & 0.537 \\
XMem\_RGBD~\cite{Xmem} &  2022   & 0.617 &0.680  & 0.649           \\
SAMTrack\_RGBD~\cite{samtrack}  &  2023   & 0.445 &0.463  & 0.454           \\
ARKitTrack~\cite{arkittrack}  &  2023   & 0.625 &0.698  & 0.662         \\
 \hline
Ours           &   -  & \textbf{0.673} &\textbf{0.723}  & \textbf{0.698}   \\ \hline
\end{tabular}
\end{center}
\end{table}

To validate the effectiveness of the proposed method, we conduct comparative experiments with seven state-of-the-art methods, including STCN~\cite{STCN}, AOT~\cite{AOT}, RPCM~\cite{RPCM}, QDMN~\cite{QDMN}, XMem~\cite{Xmem}, SAMTrack~\cite{samtrack}, and ARKitTrack~\cite{arkittrack}. In addition, we modify Xmem and SAMTrack by adding a depth branch, which is then fused with the RGB branch to form Xmem\_RGBD and SAMTrack\_RGBD. To ensure fairness in the experimental evaluation, all methods are fine-tuned on the training dataset provided by ARKitTrack ~\cite{arkittrack}.

The comparison results are presented in Table~\ref{tab:1}, illustrating that our method achieves the best performance in all three metrics. In comparison to the state-of-the-art method ARKitTrack~\cite{arkittrack}, our method exhibits improvements of 4.8\%, 2.5\%, and 3.6\% in $\mathcal{J_M}$, $\mathcal{F_M}$, and $\mathcal{J\&F}$, respectively. These advancements can be primarily attributed to the enhanced multi-store memory, which effectively extracts appearance features and establishes historical connections of segmentation results, thereby enhancing segmentation reliability significantly. When compared to baseline, XMem~\cite{Xmem}, the performance of the proposed method has even more substantial improvements with increases of 13.7\%, 6.8\%, and 14.5\% in the three metrics, respectively.
The proposed segmentation refinement module is primarily responsible for this improvement, as it effectively corrects segmentation errors and prevents their inclusion in the multi-store memory, thereby avoiding subsequent misleading segmentation. 

Compared to SAMTrack, which also utilizes the SAM for segmentation guidance, our method shows notable improvements due to the incorporation of spatio-temporal embedding and modality embedding. The spatio-temporal embedding generates accurately mixed prompts, while the modality embedding fuses informative RGB-D images with a pixel-wise strategy. In contrast, SAMTrack only performs full-image segmentation on a single modality, which limits its efficient utilization of SAM. Moreover, the HMSF extracts and fuses the complementary features from each modality, thus improving the representation ability of features, and further enhancing the performance of RGB-D segmentation. The comparison  experiments fully demonstrate the effectivenes of the proposed method.

\begin{table}[t]
\begin{center}
\caption{Ablation study on different components. HMSF represents hierarchical modality selection and fusion, STE and ME denote spatio-temporal embedding and modality embedding respectively, The best results are highlighted in \textbf{bold}.} \label{tab:2}
\begin{tabular}{cccccc}
\hline
 HMSF & STE & ME &  $\mathcal{J_M}\uparrow$  &  $\mathcal{F_M}\uparrow$   &  $\mathcal{J\&F}\uparrow$  \\ \hline
  &  &     & 0.617 &0.680  & 0.649   \\
 \checkmark  &   & & 0.637 &0.691  & 0.664   \\
 \checkmark     &  \checkmark &&  0.651 &0.702  & 0.677   \\
 \checkmark     & \checkmark & \checkmark & \textbf{0.673} &\textbf{0.723}  & \textbf{0.698}   \\ \hline
\end{tabular}
\end{center}
\end{table}
\begin{table}[t]
\begin{center}
\caption{Comparison of different thresholds, $E$ represents the threshold for regional significance and $M$ is the threshold for significant object shift, all results in the table are $\mathcal{J\&F}$. The best result is highlighted in \textbf{bold}.}
\label{tab:3}
\begin{tabular}{cccc}
\hline
 & \emph{E}=4 & \emph{E}=6 & \emph{E}=8  \\ \hline
 \emph{M}=300   & 0.673  & 0.687 & 0.654   \\
 \emph{M}=500   & 0.681  & \textbf{0.698} & 0.666   \\
 \emph{M}=700    & 0.666  & 0.767 & 0.659   \\ \hline
\end{tabular}
\end{center}
\end{table}
\subsection{Ablation Study}
\textbf{Components Analysis.} To further validate the effectiveness of each component, we conduct ablation experiments with four versions of the proposed method. 
\begin{figure*}[t]
    \centering
    \includegraphics[scale=0.7]{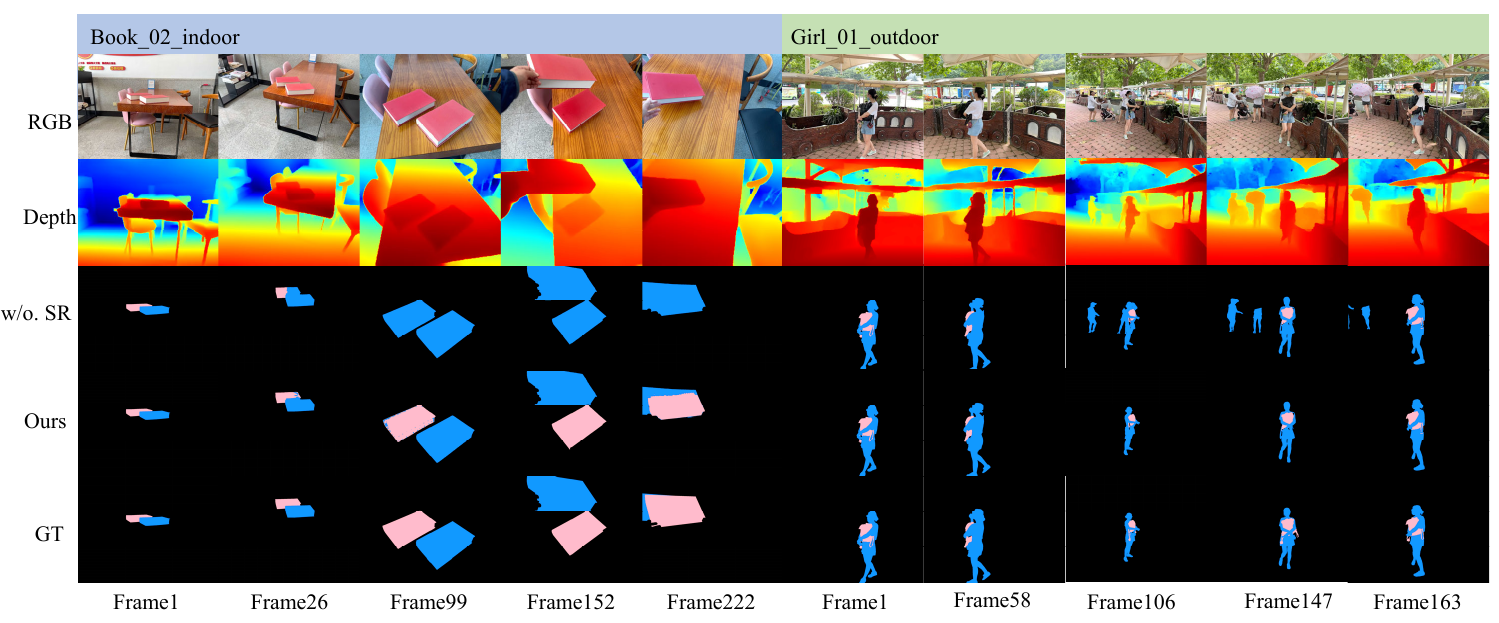}
    \caption{Qualitative comparison of the proposed method in handling different challenging scenarios, the first row shows the RGB images, the second row presents the depth images, the third row depicts the segmentation results without segmentation refinement, the fourth row illustrates the results of the proposed method, and the final row shows the ground truth. }
    \label{fig:visual}
\end{figure*} 

As shown in Table~\ref{tab:2}, in the first row, we add depth and RGB directly in the RGB-D fusion and mask generation module, as well as the multi-store memory management module. The corresponding three metrics are 0.617, 0.680, and 0.649, respectively. For the second row, instead of direct addition in RGB-D fusion mask generation and multi-store memory management module, we introduce HMSF for improved performance. The three metrics show respective increases of 2\%, 1.1\%, and 1.5\%. These results indicate that our HMSF effectively integrates dual-modality information and encodes the memory content, increasing the robustness of segmentation in complex scenarios. In the third row, we introduce a spatio-temporal embedding to generate mixed prompts, only RGB modality is fed into SAM. The metrics show improvements of 1.5\%, 1.1\%, and 1.3\%, respectively. This demonstrates that spatio-temporal embedding can enhance SAM's segmentation quality by providing reliable mixed prompts enriched with complementary dual-modalities information. The final row involves the modality embedding, which utilizes RGB-D dual-modalities for segmentation refinement. The final scores reach 0.673, 0.723, and 0.698, respectively. These results demonstrate that the modality embedding employs a pixel-wise fusion strategy based on the fused weights learned from HMSF and region significance. Thus, this process effectively provides informative segmentation images for SAM, enriched with complementary dual-modalities information. The component analysis indicates the effectiveness of each component. 

\textbf{Parameter Analysis.}
In this section, we conduct an analysis of the parameters set for the proposed method, focusing on the thresholds set for spatio-temporal embedding and modality embedding in segmentation refinement. These thresholds include the threshold for significant object shift $M$ and the threshold for regional significance $E$.

We conduct a total of nine sets of experiments, varying the value of $M$ at 300 pixels, 500 pixels, and 700 pixels, and $E$ at 4, 6, and 8, respectively. As indicated in Table~\ref{tab:3}, the best results are achieved when $M$ is set to 500 pixels and $E$ to 6. The parameter analysis suggests that the significant object shift $M$ is primarily used to estimate whether an object is segmented incorrectly based on its spatial change, hence determining whether to generate prompts from memory masks or current masks. If this value is set too low, normal object movements may be misclassified as significant shifts, leading to erroneous generation of prompts from historical information and missing the actual position of objects which can cause segmentation errors. Conversely, if the threshold is set too high, segmentation errors may be ignored thereby missing opportunities for correction.

The regional significance threshold $E$ estimates the amount of information contained in the depth modality through entropy measurement, it determines whether fusion with RGB is necessary. If this value is set too low, it may lead to the integration of potentially misleading depth information with low distinctiveness into the RGB image. On the other hand, setting it too high may disregard informative depth images containing discriminative features. The parameter analysis fully proves the rationality of parameter settings.

\textbf{Qualitative Analysis.} To demonstrate the advantages of the proposed method, we conduct a qualitative analysis, as depicted in Figure~\ref{fig:visual}. 
Given the lack of test results provided by most existing methods, we primarily employed results visualization to validate the efficacy of the segmentation refinement module.  

In the first sequence, the objects are specified as two books with confusing appearances and similar depths in the depth modality. More importantly, their depth closely resembled that of the background which posed a significant challenge for segmentation. Without segmentation refinement, our method initially segments correctly but gradually confuses these objects due to unreliable segmentation results being stored in memory leading to error accumulation and eventual complete confusion between them. 
However, with segmentation refinement utilizing spatio-temporal embedding and mixed prompt information generation through SAM upon identifying any errors during the segmentation process followed by pixel-wise fusion using modality embedding generates accurate segmentations which are then stored in memory for guiding subsequent tasks.

In the second sequence, the objects are specified as a person and her bag in the frame. The segmentation process is complicated by challenges such as the person's movements, which can occlude the backpack. Additionally, the presence of multiple similar-looking persons in the frame can lead to confusion with the segmentation object. The method without segmentation refinement might confuse one person for another. Our method leverages the segmentation refinement module to reliably segmentation and identify each object.

The qualitative analysis visually demonstrates the effectiveness of our method in complex scenarios.

\section{conclusion}
In this paper, we proposed a novel RGB-D VOS method based on an enhanced multi-store memory. Specifically, we proposed a segmentation refinement module to improve the reliability of segmentation results by incorporating spatio-temporal embedding and modality embedding which provide mixed prompts and pixel-wise fused images for SAM. Additionally, we presented the HMSF for the selective fusion of hierarchical features across both modalities, thereby enhancing the modality interaction and memory encoding. The proposed method achieved state-of-the-art performance on the latest RGB-D VOS dataset ARKitTrack. 

\begin{acks}
This work was supported by the National Natural Science Foundation of China (62072232), the Key R\&D Project of Jiangsu Province (BE2022138), the Fundamental Research Funds for the Central Universities (021714380026), and the Collaborative Innovation Center of Novel Software Technology and Industrialization.
\end{acks}
\bibliographystyle{ACM-Reference-Format}
\balance
\bibliography{sample-base}


\begin{thebibliography}{50}


\ifx \showCODEN    \undefined \def \showCODEN     #1{\unskip}     \fi
\ifx \showDOI      \undefined \def \showDOI       #1{#1}\fi
\ifx \showISBNx    \undefined \def \showISBNx     #1{\unskip}     \fi
\ifx \showISBNxiii \undefined \def \showISBNxiii  #1{\unskip}     \fi
\ifx \showISSN     \undefined \def \showISSN      #1{\unskip}     \fi
\ifx \showLCCN     \undefined \def \showLCCN      #1{\unskip}     \fi
\ifx \shownote     \undefined \def \shownote      #1{#1}          \fi
\ifx \showarticletitle \undefined \def \showarticletitle #1{#1}   \fi
\ifx \showURL      \undefined \def \showURL       {\relax}        \fi
\providecommand\bibfield[2]{#2}
\providecommand\bibinfo[2]{#2}
\providecommand\natexlab[1]{#1}
\providecommand\showeprint[2][]{arXiv:#2}

\bibitem[Bibi et~al\mbox{.}(2016)]%
        {bibi}
\bibfield{author}{\bibinfo{person}{Adel Bibi}, \bibinfo{person}{Tianzhu Zhang}, {and} \bibinfo{person}{Bernard Ghanem}.} \bibinfo{year}{2016}\natexlab{}.
\newblock \showarticletitle{3D Part-Based Sparse Tracker with Automatic Synchronization and Registration}. In \bibinfo{booktitle}{\emph{IEEE Conference on Computer Vision and Pattern Recognition}}.
\newblock


\bibitem[Caelles et~al\mbox{.}(2017)]%
        {caelles2017one}
\bibfield{author}{\bibinfo{person}{Sergi Caelles}, \bibinfo{person}{Kevis-Kokitsi Maninis}, \bibinfo{person}{Jordi Pont-Tuset}, \bibinfo{person}{Laura Leal-Taix{\'e}}, \bibinfo{person}{Daniel Cremers}, {and} \bibinfo{person}{Luc Van~Gool}.} \bibinfo{year}{2017}\natexlab{}.
\newblock \showarticletitle{One-Shot Video Object Segmentation}. In \bibinfo{booktitle}{\emph{IEEE Conference on Computer Vision and Pattern Recognition}}.
\newblock


\bibitem[Chen et~al\mbox{.}(2018)]%
        {active1}
\bibfield{author}{\bibinfo{person}{Yuhua Chen}, \bibinfo{person}{Jordi Pont-Tuset}, \bibinfo{person}{Alberto Montes}, {and} \bibinfo{person}{Luc Van~Gool}.} \bibinfo{year}{2018}\natexlab{}.
\newblock \showarticletitle{Blazingly Fast Video Object Segmentation with Pixel-wise Metric Learning}. In \bibinfo{booktitle}{\emph{IEEE Conference on Computer Vision and Pattern Recognition}}.
\newblock


\bibitem[Cheng and Schwing(2022)]%
        {Xmem}
\bibfield{author}{\bibinfo{person}{Ho~Kei Cheng} {and} \bibinfo{person}{Alexander~G Schwing}.} \bibinfo{year}{2022}\natexlab{}.
\newblock \showarticletitle{Xmem: Long-term Video Object Segmentation with an Atkinson-Shiffrin Memory Model}. In \bibinfo{booktitle}{\emph{European Conference on Computer Vision}}.
\newblock


\bibitem[Cheng et~al\mbox{.}(2021)]%
        {STCN}
\bibfield{author}{\bibinfo{person}{Ho~Kei Cheng}, \bibinfo{person}{Yu-Wing Tai}, {and} \bibinfo{person}{Chi-Keung Tang}.} \bibinfo{year}{2021}\natexlab{}.
\newblock \showarticletitle{Rethinking Space-Time Networks with Improved Memory Coverage for Efficient Video Object Segmentation}. In \bibinfo{booktitle}{\emph{Neural Information Processing Systems}}.
\newblock


\bibitem[Cheng et~al\mbox{.}(2017)]%
        {segflow}
\bibfield{author}{\bibinfo{person}{Jingchun Cheng}, \bibinfo{person}{Yi-Hsuan Tsai}, \bibinfo{person}{Shengjin Wang}, {and} \bibinfo{person}{Ming-Hsuan Yang}.} \bibinfo{year}{2017}\natexlab{}.
\newblock \showarticletitle{Segflow: Joint Learning for Video Object Segmentation and Optical Flow}. In \bibinfo{booktitle}{\emph{IEEE International Conference on Computer Vision}}.
\newblock


\bibitem[Cheng et~al\mbox{.}(2023)]%
        {samtrack}
\bibfield{author}{\bibinfo{person}{Yangming Cheng}, \bibinfo{person}{Liulei Li}, \bibinfo{person}{Yuanyou Xu}, \bibinfo{person}{Xiaodi Li}, \bibinfo{person}{Zongxin Yang}, \bibinfo{person}{Wenguan Wang}, {and} \bibinfo{person}{Yi Yang}.} \bibinfo{year}{2023}\natexlab{}.
\newblock \showarticletitle{Segment and Track Anything}.
\newblock \bibinfo{journal}{\emph{arXiv preprint arXiv:2305.06558}} (\bibinfo{year}{2023}).
\newblock


\bibitem[Dutt~Jain et~al\mbox{.}(2017)]%
        {un1}
\bibfield{author}{\bibinfo{person}{Suyog Dutt~Jain}, \bibinfo{person}{Bo Xiong}, {and} \bibinfo{person}{Kristen Grauman}.} \bibinfo{year}{2017}\natexlab{}.
\newblock \showarticletitle{Fusionseg: Learning to Combine Motion and Appearance for Fully Automatic Segmentation of Generic Objects in Videos}. In \bibinfo{booktitle}{\emph{IEEE Conference on Computer Vision and Pattern Recognition}}.
\newblock


\bibitem[Ebbinghaus(2013)]%
        {memory}
\bibfield{author}{\bibinfo{person}{Hermann Ebbinghaus}.} \bibinfo{year}{2013}\natexlab{}.
\newblock \showarticletitle{Memory: A contribution to experimental psychology}.
\newblock \bibinfo{journal}{\emph{Annals of Neurosciences}} \bibinfo{volume}{20}, \bibinfo{number}{4} (\bibinfo{year}{2013}), \bibinfo{pages}{155}.
\newblock


\bibitem[Feng and Su(2022)]%
        {rgbt1}
\bibfield{author}{\bibinfo{person}{M. Feng} {and} \bibinfo{person}{J. Su}.} \bibinfo{year}{2022}\natexlab{}.
\newblock \showarticletitle{Learning Reliable Modal Weight with Transformer for Robust RGBT Tracking}.
\newblock \bibinfo{journal}{\emph{Knowledge Based Systems}}  \bibinfo{volume}{249} (\bibinfo{year}{2022}), \bibinfo{pages}{108945}.
\newblock


\bibitem[He et~al\mbox{.}(2016)]%
        {resnet}
\bibfield{author}{\bibinfo{person}{Kaiming He}, \bibinfo{person}{Xiangyu Zhang}, \bibinfo{person}{Shaoqing Ren}, {and} \bibinfo{person}{Jian Sun}.} \bibinfo{year}{2016}\natexlab{}.
\newblock \showarticletitle{Deep Residual Learning for Image Recognition}. In \bibinfo{booktitle}{\emph{IEEE Conference on Computer Vision and Pattern Recognition}}.
\newblock


\bibitem[Hou et~al\mbox{.}(2022)]%
        {hou1}
\bibfield{author}{\bibinfo{person}{Ruichao Hou}, \bibinfo{person}{Tongwei Ren}, {and} \bibinfo{person}{Gangshan Wu}.} \bibinfo{year}{2022}\natexlab{}.
\newblock \showarticletitle{MIRNet: A Robust RGBT Tracking Jointly with Multi-Modal Interaction and Refinement}. In \bibinfo{booktitle}{\emph{IEEE International Conference on Multimedia and Expo}}.
\newblock


\bibitem[Hou et~al\mbox{.}(2023)]%
        {hou2}
\bibfield{author}{\bibinfo{person}{Ruichao Hou}, \bibinfo{person}{Boyue Xu}, \bibinfo{person}{Tongwei Ren}, {and} \bibinfo{person}{Gangshan Wu}.} \bibinfo{year}{2023}\natexlab{}.
\newblock \showarticletitle{MTNet: Learning Modality-aware Representation with Transformer for RGBT Tracking}. In \bibinfo{booktitle}{\emph{IEEE International Conference on Multimedia and Expo}}.
\newblock


\bibitem[Hu et~al\mbox{.}(2021)]%
        {hu2021learning}
\bibfield{author}{\bibinfo{person}{Li Hu}, \bibinfo{person}{Peng Zhang}, \bibinfo{person}{Bang Zhang}, \bibinfo{person}{Pan Pan}, \bibinfo{person}{Yinghui Xu}, {and} \bibinfo{person}{Rong Jin}.} \bibinfo{year}{2021}\natexlab{}.
\newblock \showarticletitle{Learning Position and Target Consistency for Memory-Based Video Object Segmentation}. In \bibinfo{booktitle}{\emph{IEEE Conference on Computer Vision and Pattern Recognition}}.
\newblock


\bibitem[Hu et~al\mbox{.}(2018)]%
        {hu2018videomatch}
\bibfield{author}{\bibinfo{person}{Yuan-Ting Hu}, \bibinfo{person}{Jia-Bin Huang}, {and} \bibinfo{person}{Alexander~G Schwing}.} \bibinfo{year}{2018}\natexlab{}.
\newblock \showarticletitle{Videomatch: Matching Based Video Object Segmentation}. In \bibinfo{booktitle}{\emph{European Conference on Computer Vision}}.
\newblock


\bibitem[Kart et~al\mbox{.}(2019)]%
        {KARTrgbd}
\bibfield{author}{\bibinfo{person}{Uur Kart}, \bibinfo{person}{Joni~Kristian Kmrinen}, {and} \bibinfo{person}{Jií Matas}.} \bibinfo{year}{2019}\natexlab{}.
\newblock \showarticletitle{How to Make an RGBD Tracker?}. In \bibinfo{booktitle}{\emph{European Conference on Computer Vision Workshop}}.
\newblock


\bibitem[Khoreva et~al\mbox{.}(2017)]%
        {lucid}
\bibfield{author}{\bibinfo{person}{Anna Khoreva}, \bibinfo{person}{Rodrigo Benenson}, \bibinfo{person}{Eddy Ilg}, \bibinfo{person}{Thomas Brox}, {and} \bibinfo{person}{Bernt Schiele}.} \bibinfo{year}{2017}\natexlab{}.
\newblock \showarticletitle{Lucid Data Dreaming for Object Tracking}. In \bibinfo{booktitle}{\emph{The DAVIS Challenge on Video Object Segmentation}}.
\newblock


\bibitem[Kirillov et~al\mbox{.}(2023)]%
        {SAM}
\bibfield{author}{\bibinfo{person}{Alexander Kirillov}, \bibinfo{person}{Eric Mintun}, \bibinfo{person}{Nikhila Ravi}, \bibinfo{person}{Hanzi Mao}, \bibinfo{person}{Chloe Rolland}, \bibinfo{person}{Laura Gustafson}, \bibinfo{person}{Tete Xiao}, \bibinfo{person}{Spencer Whitehead}, \bibinfo{person}{Alexander~C Berg}, \bibinfo{person}{Wan-Yen Lo}, {et~al\mbox{.}}} \bibinfo{year}{2023}\natexlab{}.
\newblock \showarticletitle{Segment Anything}. In \bibinfo{booktitle}{\emph{IEEE International Conference on Computer Vision}}.
\newblock


\bibitem[Li et~al\mbox{.}(2023)]%
        {drive2}
\bibfield{author}{\bibinfo{person}{Jiaqi Li}, \bibinfo{person}{Yiran Wang}, \bibinfo{person}{Zihao Huang}, \bibinfo{person}{Jinghong Zheng}, \bibinfo{person}{Ke Xian}, \bibinfo{person}{Zhiguo Cao}, {and} \bibinfo{person}{Jianming Zhang}.} \bibinfo{year}{2023}\natexlab{}.
\newblock \showarticletitle{Diffusion-Augmented Depth Prediction with Sparse Annotations}. In \bibinfo{booktitle}{\emph{ACM International Conference on Multimedia}}.
\newblock


\bibitem[Liu et~al\mbox{.}(2020)]%
        {reconst3}
\bibfield{author}{\bibinfo{person}{Caixia Liu}, \bibinfo{person}{Dehui Kong}, \bibinfo{person}{Shaofan Wang}, \bibinfo{person}{Jinghua Li}, {and} \bibinfo{person}{Baocai Yin}.} \bibinfo{year}{2020}\natexlab{}.
\newblock \showarticletitle{DLGAN: Depth-Preserving Latent Generative Adversarial Network for 3D Reconstruction}.
\newblock \bibinfo{journal}{\emph{IEEE Transactions on Multimedia}}  \bibinfo{volume}{23} (\bibinfo{year}{2020}), \bibinfo{pages}{2843--2856}.
\newblock


\bibitem[Liu et~al\mbox{.}(2022)]%
        {QDMN}
\bibfield{author}{\bibinfo{person}{Yong Liu}, \bibinfo{person}{Ran Yu}, \bibinfo{person}{Fei Yin}, \bibinfo{person}{Xinyuan Zhao}, \bibinfo{person}{Wei Zhao}, \bibinfo{person}{Weihao Xia}, {and} \bibinfo{person}{Yujiu Yang}.} \bibinfo{year}{2022}\natexlab{}.
\newblock \showarticletitle{Learning Quality-Aware Dynamic Memory for Video Object Segmentation}. In \bibinfo{booktitle}{\emph{European Conference on Computer Vision}}.
\newblock


\bibitem[Loshchilov and Hutter(2017)]%
        {adam}
\bibfield{author}{\bibinfo{person}{Ilya Loshchilov} {and} \bibinfo{person}{Frank Hutter}.} \bibinfo{year}{2017}\natexlab{}.
\newblock \showarticletitle{Decoupled Weight Decay Regularization}.
\newblock \bibinfo{journal}{\emph{arXiv preprint arXiv:1711.05101}} (\bibinfo{year}{2017}).
\newblock


\bibitem[Ma et~al\mbox{.}(2022)]%
        {drive}
\bibfield{author}{\bibinfo{person}{Zeyu Ma}, \bibinfo{person}{Yang Yang}, \bibinfo{person}{Guoqing Wang}, \bibinfo{person}{Xing Xu}, \bibinfo{person}{Heng~Tao Shen}, {and} \bibinfo{person}{Mingxing Zhang}.} \bibinfo{year}{2022}\natexlab{}.
\newblock \showarticletitle{Rethinking Open-World Object Detection in Autonomous Driving Scenarios}. In \bibinfo{booktitle}{\emph{ACM International Conference on Multimedia}}.
\newblock


\bibitem[Maninis et~al\mbox{.}(2018)]%
        {active2}
\bibfield{author}{\bibinfo{person}{Kevis-Kokitsi Maninis}, \bibinfo{person}{Sergi Caelles}, \bibinfo{person}{Jordi Pont-Tuset}, {and} \bibinfo{person}{Luc Van~Gool}.} \bibinfo{year}{2018}\natexlab{}.
\newblock \showarticletitle{Deep extreme cut: From extreme points to object segmentation}. In \bibinfo{booktitle}{\emph{IEEE Conference on Computer Vision and Pattern Recognition}}.
\newblock


\bibitem[Ouali et~al\mbox{.}(2020)]%
        {bce}
\bibfield{author}{\bibinfo{person}{Yassine Ouali}, \bibinfo{person}{C{\'e}line Hudelot}, {and} \bibinfo{person}{Myriam Tami}.} \bibinfo{year}{2020}\natexlab{}.
\newblock \showarticletitle{Semi-supervised Semantic Segmentation with Cross-Consistency Training}. In \bibinfo{booktitle}{\emph{IEEE Conference on Computer Vision and Pattern Recognition}}.
\newblock


\bibitem[Pan et~al\mbox{.}(2023)]%
        {reconst}
\bibfield{author}{\bibinfo{person}{Zihao Pan}, \bibinfo{person}{Junyi Hou}, {and} \bibinfo{person}{Lei Yu}.} \bibinfo{year}{2023}\natexlab{}.
\newblock \showarticletitle{Optimization RGB-D 3-D Reconstruction Algorithm Based on Dynamic SLAM}.
\newblock \bibinfo{journal}{\emph{IEEE Transactions on Instrumentation and Measurement}}  \bibinfo{volume}{72} (\bibinfo{year}{2023}), \bibinfo{pages}{1--13}.
\newblock


\bibitem[Qian et~al\mbox{.}(2021)]%
        {dal}
\bibfield{author}{\bibinfo{person}{Yanlin Qian}, \bibinfo{person}{Song Yan}, \bibinfo{person}{Alan Luke{\v{z}}i{\v{c}}}, \bibinfo{person}{Matej Kristan}, \bibinfo{person}{Joni-Kristian K{\"a}m{\"a}r{\"a}inen}, {and} \bibinfo{person}{Ji{\v{r}}{\'\i} Matas}.} \bibinfo{year}{2021}\natexlab{}.
\newblock \showarticletitle{DAL: A Deep Depth-aware Long-term Tracker}. In \bibinfo{booktitle}{\emph{International Conference on Pattern Recognition}}.
\newblock


\bibitem[Sharir et~al\mbox{.}(2017)]%
        {time1}
\bibfield{author}{\bibinfo{person}{Gilad Sharir}, \bibinfo{person}{Eddie Smolyansky}, {and} \bibinfo{person}{Itamar Friedman}.} \bibinfo{year}{2017}\natexlab{}.
\newblock \showarticletitle{Video Object Segmentation Using Tracked Object Proposals}.
\newblock \bibinfo{journal}{\emph{arXiv preprint arXiv:1707.06545}} (\bibinfo{year}{2017}).
\newblock


\bibitem[Shin~Yoon et~al\mbox{.}(2017)]%
        {shin2017pixel}
\bibfield{author}{\bibinfo{person}{Jae Shin~Yoon}, \bibinfo{person}{Francois Rameau}, \bibinfo{person}{Junsik Kim}, \bibinfo{person}{Seokju Lee}, \bibinfo{person}{Seunghak Shin}, {and} \bibinfo{person}{In So~Kweon}.} \bibinfo{year}{2017}\natexlab{}.
\newblock \showarticletitle{Pixel-Level Matching for Video Object Segmentation Using Convolutional Neural Networks}. In \bibinfo{booktitle}{\emph{IEEE International Conference on Computer Vision}}.
\newblock


\bibitem[Song and Xiao(2014)]%
        {PTB}
\bibfield{author}{\bibinfo{person}{S. Song} {and} \bibinfo{person}{J. Xiao}.} \bibinfo{year}{2014}\natexlab{}.
\newblock \showarticletitle{Tracking Revisited Using RGBD Camera: Unified Benchmark and Baselines}. In \bibinfo{booktitle}{\emph{IEEE International Conference on Computer Vision}}.
\newblock


\bibitem[Tang et~al\mbox{.}(2023)]%
        {fusion}
\bibfield{author}{\bibinfo{person}{Zhangyong Tang}, \bibinfo{person}{Tianyang Xu}, \bibinfo{person}{Hui Li}, \bibinfo{person}{Xiao-Jun Wu}, \bibinfo{person}{XueFeng Zhu}, {and} \bibinfo{person}{Josef Kittler}.} \bibinfo{year}{2023}\natexlab{}.
\newblock \showarticletitle{Exploring Fusion Strategies for Accurate RGBT Visual Object Tracking}.
\newblock \bibinfo{journal}{\emph{Information Fusion}}  \bibinfo{volume}{99} (\bibinfo{year}{2023}), \bibinfo{pages}{101881}.
\newblock


\bibitem[Tokmakov et~al\mbox{.}(2017)]%
        {un2}
\bibfield{author}{\bibinfo{person}{Pavel Tokmakov}, \bibinfo{person}{Karteek Alahari}, {and} \bibinfo{person}{Cordelia Schmid}.} \bibinfo{year}{2017}\natexlab{}.
\newblock \showarticletitle{Learning video object segmentation with visual memory}. In \bibinfo{booktitle}{\emph{IEEE International Conference on Computer Vision}}.
\newblock


\bibitem[Xiao et~al\mbox{.}(2023)]%
        {driving}
\bibfield{author}{\bibinfo{person}{Xiaoyang Xiao}, \bibinfo{person}{Yuqian Zhao}, \bibinfo{person}{Fan Zhang}, \bibinfo{person}{Biao Luo}, \bibinfo{person}{Lingli Yu}, \bibinfo{person}{Baifan Chen}, {and} \bibinfo{person}{Chunhua Yang}.} \bibinfo{year}{2023}\natexlab{}.
\newblock \showarticletitle{BASeg: Boundary Aware Semantic Segmentation for Autonomous Driving}.
\newblock \bibinfo{journal}{\emph{Neural Networks}}  \bibinfo{volume}{157} (\bibinfo{year}{2023}), \bibinfo{pages}{460--470}.
\newblock


\bibitem[Xiao et~al\mbox{.}(2022)]%
        {RGBT2}
\bibfield{author}{\bibinfo{person}{Yun Xiao}, \bibinfo{person}{Mengmeng Yang}, \bibinfo{person}{Chenglong Li}, \bibinfo{person}{Lei Liu}, {and} \bibinfo{person}{Jin Tang}.} \bibinfo{year}{2022}\natexlab{}.
\newblock \showarticletitle{Attribute-based Progressive Fusion Network for Rgbt Tracking}. In \bibinfo{booktitle}{\emph{AAAI Conference on Artificial Intelligence}}.
\newblock


\bibitem[Xu et~al\mbox{.}(2024)]%
        {UAV}
\bibfield{author}{\bibinfo{person}{Boyue Xu}, \bibinfo{person}{Ruichao Hou}, \bibinfo{person}{Jia Bei}, \bibinfo{person}{Tongwei Ren}, {and} \bibinfo{person}{Gangshan Wu}.} \bibinfo{year}{2024}\natexlab{}.
\newblock \showarticletitle{Jointly Modeling Association and Motion Cues for Robust Infrared UAV Tracking}.
\newblock \bibinfo{journal}{\emph{The Visual Computer}} (\bibinfo{year}{2024}), \bibinfo{pages}{1--12}.
\newblock


\bibitem[Xu et~al\mbox{.}(2023)]%
        {xu1}
\bibfield{author}{\bibinfo{person}{Boyue Xu}, \bibinfo{person}{Yi Xu}, \bibinfo{person}{Ruichao Hou}, \bibinfo{person}{Jia Bei}, \bibinfo{person}{Tongwei Ren}, {and} \bibinfo{person}{Gangshan Wu}.} \bibinfo{year}{2023}\natexlab{}.
\newblock \showarticletitle{RGB-D Tracking via Hierarchical Modality Aggregation and Distribution Network}. In \bibinfo{booktitle}{\emph{ACM International Conference on Multimedia in Asia}}.
\newblock


\bibitem[Xu et~al\mbox{.}(2018)]%
        {youtube}
\bibfield{author}{\bibinfo{person}{Ning Xu}, \bibinfo{person}{Linjie Yang}, \bibinfo{person}{Yuchen Fan}, \bibinfo{person}{Jianchao Yang}, \bibinfo{person}{Dingcheng Yue}, \bibinfo{person}{Yuchen Liang}, \bibinfo{person}{Brian Price}, \bibinfo{person}{Scott Cohen}, {and} \bibinfo{person}{Thomas Huang}.} \bibinfo{year}{2018}\natexlab{}.
\newblock \showarticletitle{Youtube-VOS: Sequence-to-Sequence Video Object Segmentation}. In \bibinfo{booktitle}{\emph{European Conference on Computer Vision}}.
\newblock


\bibitem[Xu et~al\mbox{.}(2022)]%
        {RPCM}
\bibfield{author}{\bibinfo{person}{Xiaohao Xu}, \bibinfo{person}{Jinglu Wang}, \bibinfo{person}{Xiao Li}, {and} \bibinfo{person}{Yan Lu}.} \bibinfo{year}{2022}\natexlab{}.
\newblock \showarticletitle{Reliable Propagation-Correction Modulation for Video Object Segmentation}. In \bibinfo{booktitle}{\emph{AAAI Conference on Artificial Intelligence}}.
\newblock


\bibitem[Yan et~al\mbox{.}(2021)]%
        {DepthTrack}
\bibfield{author}{\bibinfo{person}{S. Yan}, \bibinfo{person}{J. Yang}, \bibinfo{person}{J. Kpyl}, \bibinfo{person}{F. Zheng}, \bibinfo{person}{A. Leonardis}, {and} \bibinfo{person}{J.~K. Kmrinen}.} \bibinfo{year}{2021}\natexlab{}.
\newblock \showarticletitle{DepthTrack : Unveiling the Power of RGBD Tracking}. In \bibinfo{booktitle}{\emph{IEEE International Conference on Computer Vision}}.
\newblock


\bibitem[Yang et~al\mbox{.}(2022)]%
        {protrack}
\bibfield{author}{\bibinfo{person}{Jinyu Yang}, \bibinfo{person}{Zhe Li}, \bibinfo{person}{Feng Zheng}, \bibinfo{person}{Ales Leonardis}, {and} \bibinfo{person}{Jingkuan Song}.} \bibinfo{year}{2022}\natexlab{}.
\newblock \showarticletitle{Prompting for Multi-modal Tracking}. In \bibinfo{booktitle}{\emph{the ACM International Conference on Multimedia}}.
\newblock


\bibitem[Yang et~al\mbox{.}(2021)]%
        {AOT}
\bibfield{author}{\bibinfo{person}{Zongxin Yang}, \bibinfo{person}{Yunchao Wei}, {and} \bibinfo{person}{Yi Yang}.} \bibinfo{year}{2021}\natexlab{}.
\newblock \showarticletitle{Associating Objects with Transformers for Video Object Segmentation}. In \bibinfo{booktitle}{\emph{Neural Information Processing Systems}}.
\newblock


\bibitem[Yang and Yang(2022)]%
        {DEAOT}
\bibfield{author}{\bibinfo{person}{Zongxin Yang} {and} \bibinfo{person}{Yi Yang}.} \bibinfo{year}{2022}\natexlab{}.
\newblock \showarticletitle{Decoupling Features in Hierarchical Propagation for Video Object Segmentation}. In \bibinfo{booktitle}{\emph{Neural Information Processing Systems}}.
\newblock


\bibitem[Yao et~al\mbox{.}(2020)]%
        {VOS}
\bibfield{author}{\bibinfo{person}{Rui Yao}, \bibinfo{person}{Guosheng Lin}, \bibinfo{person}{Shixiong Xia}, \bibinfo{person}{Jiaqi Zhao}, {and} \bibinfo{person}{Yong Zhou}.} \bibinfo{year}{2020}\natexlab{}.
\newblock \showarticletitle{Video Object Segmentation and Tracking: A Survey}.
\newblock \bibinfo{journal}{\emph{ACM Transactions on Intelligent Systems and Technology}} \bibinfo{volume}{11}, \bibinfo{number}{4} (\bibinfo{year}{2020}), \bibinfo{pages}{1--47}.
\newblock


\bibitem[Zhang et~al\mbox{.}(2023)]%
        {reconst2}
\bibfield{author}{\bibinfo{person}{Chenyangguang Zhang}, \bibinfo{person}{Zhiqiang Lou}, \bibinfo{person}{Yan Di}, \bibinfo{person}{Federico Tombari}, {and} \bibinfo{person}{Xiangyang Ji}.} \bibinfo{year}{2023}\natexlab{}.
\newblock \showarticletitle{Sst: Real-Time End-to-End Monocular 3D Reconstruction via Sparse Spatial-Temporal Guidance}. In \bibinfo{booktitle}{\emph{IEEE International Conference on Multimedia and Expo}}.
\newblock


\bibitem[Zhang et~al\mbox{.}(2020)]%
        {tjh}
\bibfield{author}{\bibinfo{person}{Dong Zhang}, \bibinfo{person}{Hanwang Zhang}, \bibinfo{person}{Jinhui Tang}, \bibinfo{person}{Xian-Sheng Hua}, {and} \bibinfo{person}{Qianru Sun}.} \bibinfo{year}{2020}\natexlab{}.
\newblock \showarticletitle{Causal Intervention for Weakly-Supervised Semantic Segmentation}. In \bibinfo{booktitle}{\emph{Advances in Neural Information Processing Systems}}.
\newblock


\bibitem[Zhao et~al\mbox{.}(2023)]%
        {arkittrack}
\bibfield{author}{\bibinfo{person}{Haojie Zhao}, \bibinfo{person}{Junsong Chen}, \bibinfo{person}{Lijun Wang}, {and} \bibinfo{person}{Huchuan Lu}.} \bibinfo{year}{2023}\natexlab{}.
\newblock \showarticletitle{ARKitTrack: A New Diverse Dataset for Tracking Using Mobile RGB-D Data}. In \bibinfo{booktitle}{\emph{IEEE Conference on Computer Vision and Pattern Recognition}}.
\newblock


\bibitem[Zhao et~al\mbox{.}(2021)]%
        {tsdm}
\bibfield{author}{\bibinfo{person}{Pengyao Zhao}, \bibinfo{person}{Quanli Liu}, \bibinfo{person}{Wei Wang}, {and} \bibinfo{person}{Qiang Guo}.} \bibinfo{year}{2021}\natexlab{}.
\newblock \showarticletitle{TSDM: Tracking by SiamRPN++ with a Depth-refiner and a Mask-generator}. In \bibinfo{booktitle}{\emph{International Conference on Pattern Recognition}}.
\newblock


\bibitem[Zhong et~al\mbox{.}(2015)]%
        {zhongRGBD}
\bibfield{author}{\bibinfo{person}{Bineng Zhong}, \bibinfo{person}{Yingju Shen}, \bibinfo{person}{Yan Chen}, \bibinfo{person}{Weibo Xie}, \bibinfo{person}{Zhen Cui}, \bibinfo{person}{Hongbo Zhang}, \bibinfo{person}{Duansheng Chen}, \bibinfo{person}{Tian Wang}, \bibinfo{person}{Xin Liu}, \bibinfo{person}{Shujuan Peng}, {et~al\mbox{.}}} \bibinfo{year}{2015}\natexlab{}.
\newblock \showarticletitle{Online Learning 3D Context for Robust Visual Tracking}.
\newblock \bibinfo{journal}{\emph{Neurocomputing}} (\bibinfo{year}{2015}).
\newblock


\bibitem[Zhu et~al\mbox{.}(2023a)]%
        {VIPT}
\bibfield{author}{\bibinfo{person}{Jiawen Zhu}, \bibinfo{person}{Simiao Lai}, \bibinfo{person}{Xin Chen}, \bibinfo{person}{Dong Wang}, {and} \bibinfo{person}{Huchuan Lu}.} \bibinfo{year}{2023}\natexlab{a}.
\newblock \showarticletitle{Visual Prompt Multi-Modal Tracking}. In \bibinfo{booktitle}{\emph{IEEE Conference on Computer Vision and Pattern Recognition}}.
\newblock


\bibitem[Zhu et~al\mbox{.}(2023b)]%
        {rgbd1k}
\bibfield{author}{\bibinfo{person}{Xue-Feng Zhu}, \bibinfo{person}{Tianyang Xu}, \bibinfo{person}{Zhangyong Tang}, \bibinfo{person}{Zucheng Wu}, \bibinfo{person}{Haodong Liu}, \bibinfo{person}{Xiao Yang}, {et~al\mbox{.}}} \bibinfo{year}{2023}\natexlab{b}.
\newblock \showarticletitle{RGBD1K: A Large-scale Dataset and Benchmark for RGB-D Object Tracking}. In \bibinfo{booktitle}{\emph{AAAI Conference on Artificial Intelligence}}.
\newblock


\end{thebibliography}
\end{document}